\documentclass[11pt]{article}
\usepackage{coling2016}
\usepackage{times}
\usepackage{url}
\usepackage{latexsym}
\usepackage{tikz}
\usepackage{standalone}
\usepackage{amsmath}
\usepackage{multirow}
\usepackage{array}
\usepackage{hhline}
\usepackage{subfigure}
\newcolumntype{P}[1]{>{\centering\arraybackslash}p{#1}}

\title{Semi Supervised Preposition-Sense Disambiguation\\ using Multilingual Data}

\author{Hila Gonen \\
  Department of Computer Science \\
  Bar-Ilan University \\
  {\tt hilagnn@gmail.com} \\\And
  Yoav Goldberg \\
  Department of Computer Science \\
  Bar-Ilan University \\
  {\tt yoav.goldberg@gmail.com} \\}

\date{}

\begin{document}
	\maketitle
	\begin{abstract}
		Prepositions are very common and very ambiguous, and understanding their sense is critical for understanding the meaning of the sentence. Supervised corpora for the preposition-sense disambiguation task are small, suggesting a semi-supervised approach to the task. We show that signals from unannotated multilingual data can be used to improve supervised preposition-sense disambiguation. Our approach pre-trains an LSTM encoder for predicting the translation of a preposition, and then incorporates the pre-trained encoder as a component in a supervised classification system, and fine-tunes it for the task. The multilingual signals consistently improve results on two preposition-sense datasets.
	\end{abstract}

\section{Introduction}

\blfootnote{ 
	
	\hspace{-0.65cm}  

	This work is licensed under a Creative Commons 
	Attribution-ShareAlike 4.0 International License.
	License details:\\
	\url{http://creativecommons.org/licenses/by-sa/4.0/}

}

Preposition-sense disambiguation \cite{LH05,LH07,SSHP15,SSH16}, is the task of assigning a category to a preposition in context (see Section \ref{psd}). Choosing the correct sense of a preposition is crucial for understanding the meaning of the text. This important semantic task is especially challenging from a learning perspective as only little amounts of annotated training data are available for it. Indeed, previous systems (see Sections \ref{psd:prev} and \ref{psd:compare}) make extensive use of the vast and human-curated WordNet lexicon \cite{M95} in order to compensate for the small size of the annotated data and obtain good accuracies. 

Instead, we propose to deal with the scarcity of annotated data by taking a semi-supervised approach.
We rely on the intuition that word ambiguity tends to differ between languages \cite{DIS91}, and show that multilingual corpora can provide a good signal for the preposition sense disambiguation task. Multilingual corpora are vast and relatively easy to obtain \cite{RS03,K05,SPW06}, making them appealing candidates for use in a semi-supervised setting.

Our approach (Section \ref{approach}) is based on representation learning \cite{BCV13}, and can also be seen as an instance of multi-task \cite{C98}, or transfer learning \cite{PY10}. First, we train an LSTM-based neural network \cite{HS97} to predict a foreign (say, French) preposition given the context of an English preposition. This trains the network to map contexts of English prepositions to representations that are predictive of corresponding foreign prepositions, which are in turn correlated with preposition senses. The learned mapper, which takes into account large amounts of parallel text, is then incorporated into a monolingual preposition-sense disambiguation system (Section \ref{model_wo}) and is fine-tuned based on the small amounts of available supervised data.  We show that the multilingual signal is effective for the preposition-sense disambiguation task on two different datasets (Section \ref{eval}).

\section{Background}

\subsection{Preposition Sense Disambiguation}
\label{psd}

Prepositions are very common, very ambiguous and tend to carry different meanings in different contexts. Consider the following 3 sentences: ``You should book a room \textit{for} 2 nights", ``\textit{For} some reason, he is not here yet" and ``I went there to get a present \textit{for} my mother". The preposition ``for" has 3 different readings in these sentences: in the first sentence it indicates \textsc{duration}, in the second it indicates an \textsc{explanation}, and in the third a \textsc{beneficiary}. The preposition-sense disambiguation task is defined as follows: given a preposition within a sentential context, decide which category it belongs to, or what its role in the sentence is. Choosing the right sense of a preposition is central to understanding the meaning of an utterance \cite{BKV09}. 

\subsubsection{Previous Work and Available Corpora}
\label{psd:prev}

The preposition-sense disambiguation task was the focus of the SemEval 2007 shared task \cite{LH07}, based on the set of senses defined in The Preposition Project (TPP) \cite{LH05}, with three participating systems \cite{YB07,Y07,PTP07}. Since then, it was tackled in several additional works \cite{DNS09,HT09,HTH10,T11,SR13b}, some of which used different preposition sense inventories and corpora, based on subsets of the TPP dictionary. Srikumar and Roth \shortcite{SR13b} modeled semantic relations expressed by prepositions. For this task, they presented a variation of the TPP inventory, by collapsing related preposition senses, so that all senses are shared between all prepositions \cite{SR13a}. Schneider et al \shortcite{SSHP15} further improve this inventory and define a new annotation scheme. 

There are two main datasets for this task: the corpus of the SemEval 2007 shared task \cite{LH07}, and the Web-reviews corpus \cite{SSH16}: 

\paragraph{SemEval 2007 Corpus}
This corpus covers 34 prepositions with 16,557 training and 8096 test sentences, each containing a single preposition example. The sentences were extracted from the FrameNet database,\footnote{\url{http://framenet.icsi.berkeley.edu/}} based mostly on the British National Corpus (with 75\%/25\% of informative-writings/literary). Each preposition has a different set of possible senses, with a range of 2 to 25 possible senses for a given preposition. We use the original split to train and test sets.

\paragraph{Web-reviews Corpus}

Schneider et al \shortcite{SSHP15} introduce a new, unified and improved sense inventory and corpus \cite{SSH16} in which all prepositions share the same set of senses (senses from a unified inventory are often referred to as supersenses). This corpus contains text in the online reviews genre. It is much smaller than the SemEval corpus, with 4,250 preposition mentions covering 114 different prepositions which are annotated into 63 fine-grained senses. The senses are grouped in a hierarchy, from which we chose a coarse-grained subset of 12 senses for this work: \textsc{Affector}, \textsc{Attribute}, \textsc{Circumstance}, \textsc{Co-Participant}, \textsc{Configuration}, \textsc{Experiencer}, \textsc{Explanation}, \textsc{Manner}, \textsc{Place}, \textsc{Stimulus}, \textsc{Temporal}, \textsc{Undergoer}.
We find the Web-reviews corpus more appealing than the SemEval one: the unified sense inventory makes the sense-predictions more suitable for use in downstream applications. 
While our focus in this work is the Web-reviews corpus, we are the first to report results on this dataset. For the sake of comparison to previous work, we also evaluate our models on the SemEval corpus.

\subsection{Neural Networks and Notation}

We use $w_{1:n}$ to indicate a list of vectors, and $w_{n:1}$ to indicate the reversed list. We use $\circ$ for vector concatenation, and $x[j]$ for selecting the $j^{th}$ element in a vector $x$.

A multi-layer perceptron (MLP) is a non linear classifier. In this work, we focus on MLPs with a single hidden layer and a softmax output transformation, and define the function $MLP(x)$ as:
$$MLP(x) = \mathrm{softmax}(U(\mathrm{g}(Wx + b1)) + b2)$$
where $g$ is a non-linear activation function such as $ReLU$ or $tanh$, $W$ and $U$ are input-to-hidden and hidden-to-output transformation matrices, and $b1$ and $b2$ are optional bias terms. We use subscripts ($MLP_{f1}$, $MLP_{f2}$) to denote MLPs with different parameters.

Recurrent Neural Networks (RNNs) \cite{E90} allow the representation of arbitrary sized sequences, without limiting the length of the history. RNN models have been proven to effectively model sequence-related phenomena such as line lengths, brackets and quotes \cite{KJF15}. 

In our implementation we use the long short-term memory network (LSTM), a subtype of the RNN \cite{HS97}. $LSTM(w_{1:i})$ is the output vector resulting from inputing the items $w_1,...,w_i$ into the LSTM in order. 

\section{Monolingual Preposition Sense Classification}
\label{model_wo}

We start by describing an MLP-based model for classifying prepositions to their senses. For an English sentence $s=w_1,...,w_n$ and a preposition position $i$,\footnote{We also support multi-word prepositions in this work. The extension is trivial. } we classify to the sense $y$ as: 
\[
y = \mathop{\mathrm{argmax}}_jMLP_{sense}(\phi(s, i))[j]
\]	
where $\phi(s,i)$ is a feature vector composed of 19 features. The features are based on the features of Tratz and Hovy \shortcite{HT09}, and are similar in spirit to those used in previous attempts at preposition sense disambiguation. We deliberately do not include WordNet based features, as we want to focus on features that do not require extensive human-curated resources. This makes our model applicable for use in other languages with minimal change. We use the following features: (1) The embedding of the preposition. (2) The embeddings of the lemmas of the two words before and after the preposition, of the head of the preposition in the dependency tree, and of the first modifier of the preposition. (3) The embeddings of the POS tags of these words, of the preposition, and of the head's head. (4) The embeddings of the labels of the edges to the head of the preposition, to the head's head and to the first modifier of the preposition. (5) A boolean that indicates whether one of the two words that follow the preposition is capitalized. The English sentences were parsed using the spaCy parser.\footnote{\url{https://spacy.io/}}  

The network (including the embedding vectors) is trained using cross entropy loss. This model performs relatively well, achieving an accuracy of 73.34 on the Web-reviews corpus, way above the most-frequent-sense baseline of 62.37. On the SemEval corpus, it achieves an accuracy of 74.8, outperforming all participants in the original shared task (Section~\ref{eval}). However, these results are limited by the small size of both training sets. In what follows, we will improve the model using unannotated data. 

\section{Semi-Supervised Learning Using Multilingual Data}
\label{approach}

Our goal is to derive a representation from unannotated data that is predictive of preposition-senses. We suggest using multilingual data, following the intuition that preposition ambiguity usually differs between languages \cite{DIS91}. For example, consider the following two sentences, taken from the Europarl parallel corpus \cite{K05}: ``What action will it take to defuse the crisis and tension \textit{in} the region?'', and ``These are only available \textit{in} English, which is totally unacceptable''. In the first sentence, the preposition ``in'' is translated into the French preposition ``dans'', whereas in the second one, it is translated into the French preposition ``en''. Thus, a representation that is predictive of the preposition's translation is likely to be predictive also of its sense.

\paragraph{Learning a representation from a multilingual corpus}

We train a neural network model to encode the context of an English preposition as a vector, and predict the foreign preposition based on the context vector. The resulting context encodings will then be predictive of the foreign prepositions, and hopefully also of the preposition senses.

We derive a training set of roughly 7.4M instances from the Europarl corpus \cite{K05}.
Europoarl contains sentence-aligned data in 21 languages. We started by using several ones, and ended up with a subset of 12 languages\footnote{Bulgarian, Czech, Danish, German, Greek, Spanish, French, Hungarian, Italian, Polish, Romanian and Swedish.} that together constitute a good representation of the different language families available in the corpus. Though adding the other languages is possible, we did not experiment with them. To extract the training set, we first word-align\footnote{Word-alignment is done using the \texttt{cdec} aligner \cite{cdec}.} the sentence-aligned data, and then create a dataset of English sentences where each preposition is matched to its translation in a foreign language. Since the alignment of prepositions is noisier than that of content words, we use a heuristic to improve precision: given a candidate foreign-preposition, we verify that the two words surrounding it are aligned to the two words surrounding the English preposition. Additionally, we filter out, for each English preposition, all foreign prepositions that were aligned to it in less than 5\% of the cases.

We then train the context representations according to the following model.
For an English sentence $s=w_1,...,w_n$, a preposition position $i$ and a target preposition $p$ in language $L$, we encode the context as a concatenation of two LSTMs, one reading the sentence from the beginning up to but not including the preposition, and the other in reverse:
\[
ctx(s,i) = LSTM_f(w_{1:i-1})\circ LSTM_b(w_{n:i+1})
\]

This is similar to a BiLSTM encoder, with the difference that the encoding does not include the preposition $w_i$ but only its context. By ignoring the preposition, we force the model to focus on the context, and help it share information between different prepositions. Indeed, including the preposition in the encoder resulted in better performance in foreign preposition classification, but the resulting representation was not as effective when used for the sense disambiguation task.  

The context vector is then fed into a language specific MLP for predicting the target preposition:

\[
\hat{p} = \mathop{\mathrm{argmax}}_jMLP_L(ctx(s,i))[j]
\]

The context-encoder and the word embeddings are shared across languages, but the MLP classifiers that follow are language specific. By using multiple languages, we learn more robust representations.

The English word embeddings can be initialized randomly, or using pre-trained embedding vectors, as we explore in Section~\ref{embeddings}. The network is trained using cross entropy loss, and the error is back-propagated through the context-encoder and the word embeddings.

\paragraph{Using the representation for sense classification}
Once the encoder is trained over the multilingual data, we incorporate it in the supervised sense-disambiguation model by concatenating the representation obtained from the context encoder to the feature vector. Concretely, the supervised model now becomes:
\[
y = \mathop{\mathrm{argmax}}_jMLP_{sense}(ctx(s, i) \circ \phi(s,i) )[j]
\]
where $ctx(s, i)$ is the output vector of the context-encoder and $\phi(s,i)$ is the feature vector as before.

\begin{figure*}[t] 
	\centering
	\scalebox{.5}{\input{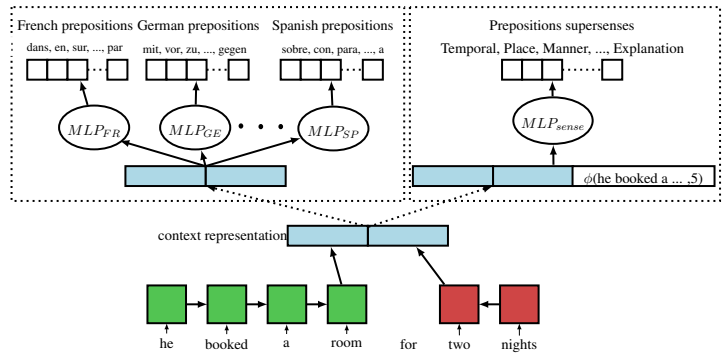}}
	\vspace*{-0.2cm}
	\caption{\footnotesize The suggested model for incorporating multilingual data in classifying prepositions to senses. First, a context-encoder (at the bottom, the green and red squares are LSTM cells) is trained on the Europarl corpus, with a different MLP for each language (left dashed frame). Then, the representation obtained from the context-encoder is added to the feature vector when classifying a preposition to senses (right dashed frame).}
	\label{fig:model}
\end{figure*}

The network is trained using cross entropy loss, and the  error back-propagates also to the context-encoder and to the word embeddings to maximize the model's ability to adapt to the preposition-sense disambiguation task. The complete model is depicted in Figure~\ref{fig:model}.

\section{Empirical results}
\label{eval}

\paragraph{Implementation details}
The models were implemented using PyCNN.\footnote{\url{https://github.com/clab/cnn}} All models were trained using SGD, shuffling the examples before each of the 5 epochs. When training a sense prediction model, we use early stopping and choose the best performing model on the development set. The sense-prediction MLP uses $ReLU$ activation, and foreign preposition MLPs use $tanh$, with no bias terms. Unless noted otherwise, we use randomly initialized embedding vectors. For each experiment, we chose the parameters that maximized the accuracy on the dev set.\footnote{In most of the experiments, the best results are achieved when the hidden-layer of the sense-prediction MLPs is of the size 500, and the preposition embedding is of size 200. In some cases, the best results are achieved with different dimensions. These two parameters were tuned on the dev set. 
The embeddings of the features are of dimension 4, with the exception of the lemmas, which are of dimension 50. The dimension of the input to the LSTMs (word embeddings) is 128. Both LSTMs have a single layer with 100 nodes, thus, the representation of the context obtained from the context-encoder is of dimension 200. The hidden-layer of the foreign-preposition MLP is of size 32.} The accuracies we report are the average accuracies over 5 different seeds.

\subsection{Evaluation on the Web-reviews corpus}

\paragraph{Using multilingual data}

Our main motivation in this work was to train a representation which is useful for the preposition-sense disambiguation task. Thus, we compare the performance of our model using the representation obtained from the context-encoder (multilingual model) with the model that does not use this representation (base model). We use the train/test split provided with the corpus. We further split the train set into train and dev sets, by assigning every fourth example of each sense to the dev set, yielding 2552/845/853 instances of train/dev/test.

The results are presented in Table~\ref{tab:main}. We see an improvement of 2.86 points when using the pre-trained context representations, improving the average result from 73.34 to 76.20.

To verify that the improvement stems from pre-training the context-encoder on multilingual data and not from adding the context-encoder as is, we also evaluated the performance of a model identical to the multilingual model, but with no pre-training on the multilingual data (context model, middle row of Table~\ref{tab:main}). The context model achieved a very similar result to that of the base model -- 73.76, indicating that adding the context-encoder to the base model is not the source of the improvement.

\begin{table}[h!]
	\centering
	\scalebox{0.7}{
		\begin{tabular}	{| p{3.6cm} | P{3.2cm} |}
			\hline
			Model & Accuracy \\ \hline
			base & 73.34 \small (71.63-73.97) \\ \hline   
			+context & 73.76 \small(71.86-75.38) \\ \hline 
			+context(multilingual) & \textbf{76.20}  \small(74.91-77.26)  \\ \hline
		\end{tabular}
	}
	\vspace*{-0.2cm}
	\caption{\footnotesize The average accuracies on the test set of the Web-reviews corpus on 5 different seeds. Numbers in brackets indicate the min and max accuracy across seeds.} 
	\label{tab:main}
	\vspace*{-0.2cm}
\end{table}

\paragraph{Using monolingual or bilingual data only}

In order to verify the contribution of incorporating information from 12 languages, we also experiment with monolingual and bilingual models. For the monolingual model we train a model similar to our multilingual one, but when trying to predict the English preposition itself, rather than the foreign one, ignoring the multilingual signal altogether. For the bilingual models we train 12 separate models similar to our multilingual model, where each one is trained only on the training examples of a single language.

As shown in Table \ref{tab:per_lang}, both the monolingual and the bilingual models improve over the base model (with the exception of Czech), but no improvement is as significant as that of the multilingual model. In addition, we see that the strength of the model does not depend solely on the number of training examples.

\begin{table}[!h]
	\centering
	\scalebox{0.7}{
		\begin{tabular}	{| p{3.8cm} | P{3.2cm} | P{2.2cm} | P{4.2cm} |}
			\hline
			Language & Accuracy & Improvement & Num. of training examples \\ \hline
			None (base model) & 73.34 \small (71.63-73.97) & -- & -- \\ \hhline{|=|=|=|=|}
			Czech & 73.06 \small(72.57-73.86) & -0.28 & 190,850 \\ \hline
			Polish & 73.93 \small(73.15-74.79) & +0.59 & 166,101 \\ \hline
			Italian & 73.97 \small (72.22-75.26) & +0.63 & 810,589 \\ \hline   
			Romanian & 74.09 \small(73.15-74.56) & +0.75 & 205,520 \\ \hline
			Hungarian & 74.42 \small(73.27-75.15) & +1.08 & 40,302 \\ \hline
			Bulgarian & 74.44 \small(73.27-74.91) & +1.10 & 292,908 \\ \hline
			Spanish & 74.65 \small(73.51-75.73) & +1.31 & 1,267,400 \\ \hline
			German & 74.73 \small(73.74-75.62) & +1.39 & 603,861 \\ \hline
			Danish & 75.08 \small(74.21-77.49) & +1.74 & 1,131,915 \\ \hline
			Greek & 75.12 \small(74.09-76.20) & +1.78 & 586,494 \\ \hline
			French & 75.43 \small(74.21-77.02) & +2.09 & 1,033,267\\ \hline			
			English (monolingual) & 75.68 \small(74.79-76.55) & +2.34 & 7,483,206\\ \hline			
			Swedish & 75.87 \small(74.68-77.49) & +2.53 & 1,153,999 \\ \hhline{|=|=|=|=|}	
			All 12 languages & \textbf{76.20} \small(74.91-77.26) & \textbf{+2.86} & 7,483,206\\ \hline	
		\end{tabular}
	}
	\vspace*{-0.2cm}
	\caption{\footnotesize The average accuracies on the test set of the Web-reviews corpus on 5 different seeds, using monolingual and bilingual models, along with the improvement over the base model and the number of training examples in each language. Numbers in brackets indicate the min and max accuracy across seeds.} 
	\label{tab:per_lang}
	\vspace*{-0.2cm}
\end{table}

\paragraph{Adding external word embeddings}
\label{embeddings}

Another way of incorporating semi-supervised data into a model is using pre-trained word embeddings. We evaluate our model when using external word embeddings instead of randomly initialized word embeddings. We perform three experiments: 1. using external word embeddings only for the words that are fed into the context-encoder. 2. using external word embeddings only for the lemmas of the features. 3. using external word embeddings for both.

We use two sets of word embeddings: 5-window-bag-of-words-based and dependency-based, both trained by Levy and Goldberg \shortcite{LG14} on English Wikipedia.\footnote{\url{https://levyomer.wordpress.com/2014/04/25/dependency-based-word-embeddings/}} As shown in Table \ref{tab:allembeddings}, both pre-trained embeddings improve the performance of all models in most cases. In all cases, the multilingual model outperforms the base model and the context model, both achieving similar results. Using external word embeddings for both the features and the context-encoder helps the most. The best result of 78.55 is achieved by the multilingual model, improving the result of the base model under the same conditions by 1.71 points.

\begin{table}[!h]
	\begin{center}
		\scalebox{0.7}{
			\begin{tabular} { | p{3.4cm} | P{2.7cm} | P{2.7cm} | P{2.7cm} | P{2.7cm}| P{2.7cm} | P{2.7cm} |}
				\hline
				\centering \multirow{2}{*}{Model} & \multicolumn{2}{c|}{Context-encoder embeddings only} & \multicolumn{2}{c|}{Feature embeddings only} & 
				\multicolumn{2}{c|}{Embeddings for both} 
				\\ \cline{2-7}
				& Bow & Deps & Bow & Deps & Bow & Deps \\ \hline 
				base 
				& 73.34 \small (71.63-73.97)
				& 73.34 \small (71.63-73.97) 
				& 76.95 \small (75.85-77.96)
				& 76.84 \small(76.32-77.26)
				& 76.95 \small (75.85-77.96)
				& 76.84 \small(76.32-77.26) \\ \hline
				+context 
				& 74.07 \small(72.10-75.15)
				& 74.42 \small(73.62-75.03) 
				& 76.72 \small(75.85-77.96)
				& 77.47 \small(75.85-78.55)
				& 77.14 \small(76.79-78.08)
				& 77.73 \small(77.14-78.43) \\ \hline
				+context(multilingual) 
				& 75.57 \small(73.51-77.84) 
				& 75.90 \small(75.03-76.55)
				& 77.58 \small(77.02-78.08)
				& 77.58 \small(77.14-78.66)
				& 78.45 \small(77.49-79.48)
				& \textbf{78.55} \small(77.37-79.37) \\ \hline
			\end{tabular}
		}
		\vspace*{-0.2cm}
		\caption{\footnotesize The average accuracies on the test set of the Web-reviews corpus with different pre-trained embeddings on 5 different seeds. Numbers in brackets indicate the min and max accuracy across seeds. \textbf{Bow}: 5-words window; \textbf{Deps}: dependency-based.} 
		\vspace*{-0.2cm}
		\label{tab:allembeddings}
	\end{center}
\end{table}

\subsection{Evaluation on the SemEval corpus}

\paragraph{Adaptations to the SemEval corpus}

In the SemEval corpus each preposition has a different set of senses, and the natural approach is to learn a different model for each one. We call this the \textit{disjoint} approach. However, we found this approach a bit wasteful in terms of exploiting the annotated data, and we propose a model that uses the information from all prepositions simultaneously (\textit{unified}). In the unified approach, 
we create an MLP classifier for each preposition, but all of them share a single input-to-hidden transformation matrix and a single bias term. Formally, for a preposition $p$, we define its MLP as follows:

$$MLP_p(x) = \mathrm{softmax}(U_p(\mathrm{g}(Wx + b1)) + b2_p)$$
where $W$ is the shared input-to-hidden transformation matrix, $b1$ is the shared bias term, and $U_p$ and $b2_p$ are preposition-specific hidden-to-output transformation matrix and bias term, respectively. This unified model is trained over the training examples of all prepositions together.

The SemEval corpus sometimes provides multiple senses for a given preposition instance. In both the disjoint and the unified approaches we treat these cases by generalizing the cross entropy loss for multiple correct classes. In the common case, where each training example has a single correct class, the cross entropy loss is defined as $-\log p_i$, where $p_i$ is the probability that the model assigns to the correct class. Here, instead of using $-\log p_i$, we use $-\log(\sum_{i \in C} p_i)$, where $C$ is the set of correct classes.

\paragraph{Results} The model performs well also on the SemEval corpus, achieving an accuracy of 76.9. Note that we use the exact same parameters that were tuned on the dev set of the Web-reviews corpus, with no additional tuning on this corpus.

As shown in Table~\ref{tab:semevalseparate}, the unified model, which trains on all prepositions simultaneously, performs better than a separate model for each preposition (disjoint model), and achieves an improvement of 1.3 points when using the multilingual model. In addition, in both cases we get a significant improvement over the base model when using the pre-trained context-representation. In the unified model, adding the pre-trained context-representation improves the result by 2.1 points. As in the case of the Web-reviews corpus, we can see that this improvement does not stem from adding the context representation as is. Pre-training the representation is essential for achieving these improved results.

\begin{table}[!h]
	\begin{center}
		\scalebox{0.7}{
		\begin{tabular} { | p{3.4cm} | P{2.7cm} | P{2.7cm} |}
			\hline
			Model & Disjoint & Unified  \\ \hline 
			base & 73.7 \small(73.3-74.1) &  74.8 \small(74.4-75.4)	  \\ \hline
			+context & 73.8 \small(73.6-74.0) & 75.4 \small(74.8-75.8) 	 \\ \hline
			+context(multilingual) & 75.6 \small(75.4-75.8) & \textbf{76.9} \small(76.4-77.7)  \\ \hline
		\end{tabular}
		}
		\vspace*{-0.2cm}
		\caption{\footnotesize The average accuracies on the test set of the SemEval corpus on 5 different seeds, with both the disjoint and the unified models. Numbers in brackets indicate the min and max accuracy across seeds.} 
		\vspace*{-0.2cm}
		\label{tab:semevalseparate}
	\end{center}
\end{table}

Similar to the results on the Web-reviews Corpus, when using external word embeddings both for the words that are fed into the context-encoder and for the features, we get an improvement in all models, with an average improvement of 3 points when using the 5-words-window based embeddings. The best result amongst the three models is of 79.6 and is achieved by the multilingual model, improving over the base model by 2.5 points. The results are shown in Table~\ref{tab:semevalembed}.

Note that unlike previous experiments, adding external word embeddings improves the context model over the base model significantly, approaching the results of the multilingual model. For this reason, we also evaluated a model in which we concatenate both contexts: that of the context model (no pre-training), and that of the multilingual model (pre-trained on the multilingual data). In the case where both models achieve similar results, combining both contexts further improves the result, which indicates that they are complementary. The best result of 80.0 is achieved when using both contexts with the 5-window-bag-of-words-based embeddings. We also evaluated this combined model on the Web-reviews corpus, but got no improvement in most cases. This was predictable since in all experiments on that corpus we had a large difference between the results of the context model and of the multilingual model. The only case where we saw an improvement with both contexts was when using dependency-based embeddings for both the features and the context-encoder. The difference between the two datasets can be explained by the much larger size of the SemEval dataset, which allows the context encoder to learn from more data, even without pre-training on multilingual data.

\begin{table}[!h]
	\begin{center}
		\scalebox{0.7}{
		\begin{tabular} { | p{3.4cm} | P{2.7cm} | P{2.7cm} || P{2.7cm} |}
			\hline
			Model & Bow & Deps & None  \\ \hline 
			base & 77.1 \small(76.9-77.2)  & 76.6 \small(76.3-76.9) & 74.8 \small(74.4-75.4)	  \\ \hline
			+context & 79.5 \small(78.8-79.9)   & 78.5 \small(78.0-78.8) & 75.4 \small(74.8-75.8)	 \\ \hline
			+context(multilingual) & 79.6 \small(79.3-79.9) & 79.3 \small(78.8-79.6) & 76.9 \small(76.4-77.7)  \\ \hline
			+both contexts & \textbf{80.0} \small(79.8-80.2) & 79.2 \small(78.6-79.5) & 77.3 \small(77.2-77.5) \\ \hline
		\end{tabular}
		}
		\vspace*{-0.2cm}
		\caption{\footnotesize The average accuracies on the test set of the SemEval corpus on 5 different seeds, with the unified model, when using external word embeddings for both the context-encoder and the features. Numbers in brackets indicate the min and max accuracy across seeds. \textbf{Bow}: 5-words window; \textbf{Deps}: dependency-based; \textbf{None}: no external word embeddings.}
		\vspace*{-0.2cm}
		\label{tab:semevalembed}
	\end{center}
\end{table}

\subsection{Using Ensembles}

We create an ensemble by training 5 different models (each with a different random seed), and predict test instances using a majority vote over the models. The results are presented in Table~\ref{tab:reviewsensemble}. As expected, results in all models further improve when using the ensemble. Using the multilingual context helps also when using the ensemble. We see an improvements of 1.99 points on the web-reviews corpus, improving the result to 80.54. The performance on the SemEval corpus improves by 1.7 points, and reaches an accuracy of 81.7. These results are higher than those of the base model by 2.93 and 2.2 points, respectively.

\begin{table}[!h]
	\begin{center}
		\scalebox{0.7}{
		\begin{tabular} { | p{3.4cm} | P{3.1cm} | P{1.5cm} | P{3.1cm} | P{1.5cm}|}
			\hline
			\centering \multirow{2}{*}{Model} & \multicolumn{2}{c|}{Web-reviews Corpus} & \multicolumn{2}{c|}{SemEval Corpus}   \\ \cline{2-5}
			& Average & Ensemble & Average & Ensemble \\ \hline 
			base & 76.84 \small(76.32-77.26) & 	77.61
			& 77.1 \small(76.9-77.2)  &  79.5 \\ \hline
			+context & 77.73 \small(77.14-78.43) & 78.90 
			& 79.5 \small(78.8-79.9) & 81.1 \\ \hline
			+context(multilingual) & 78.55 \small(77.37-79.37) & \textbf{80.54} 
			& 79.6 \small(79.3-79.9) & 81.2 \\ \hline
			+both contexts & 79.34 \small(78.43-80.19) & 79.84 & 80.0 \small(79.8-80.2) & \textbf{81.7} \\ \hline
		\end{tabular}
		}
		\vspace*{-0.2cm}
		\caption{\footnotesize The results on both datasets on 5 different seeds as reported in Tables~\ref{tab:allembeddings} and~\ref{tab:semevalembed} in comparison to the results using the ensemble. Numbers in brackets indicate the min and max accuracy across seeds.} 
		\vspace*{-0.2cm}
		\label{tab:reviewsensemble}
	\end{center}
\end{table}

\vspace{-5pt}
\subsection{Comparison to previous systems}
\label{psd:compare}

Table~\ref{tab:all_semevals} compares our SemEval results with those of previous systems. The system of Ye and Baldwin \shortcite{YB07} got the highest result out of the three participating systems in the SemEval 2007 shared task. They extracted features such as POS tags and WordNet-based features, and also high level features (e.g semantic role tags), using a word window of up to seven words, in a Maximum Entropy classifier. Tratz and Hovy \shortcite{HT09} got a higher result with similar features by using a set of  positions that are syntactically related to the preposition instead of a fixed window size. The best performing systems are of Hovy et al \shortcite{HTH10} and of Srikumar and Roth \shortcite{SR13b}. Both systems rely on vast and thoroughly-engineered feature sets, including many WordNet based features. Hovy et al \shortcite{HTH10} explored different word choices (i.e, a fixed window vs. syntactically related words) and different methods of extracting them, while Srikumar and Roth \shortcite{SR13b} improved performance by jointly predicting preposition senses and relations. 

In contrast, our models do not include any WordNet based features, making them applicable also for languages lacking such resources. Our models achieve competitive results, outperforming most previous systems, despite using relatively few features and performing hyper-parameter tuning only on the different domain Web-reviews corpus. 

\begin{table}[!h]
	\centering
	\scalebox{0.7}{
	\begin{tabular}	{| p{9.6cm} | P{3.2cm} |}
		\hline
		Model & Accuracy \\ \hline
		base & 74.8 \\ \hline
		+context & 75.4  \\ \hline
	    +context(multilingual) &  76.9 \\ \hline
	    +context(multilingual) + embeddings  & 79.6 \\ \hline
	    +both contexts + embeddings & 80.0 \\ \hline
	    +both contexts + embeddings + ensemble  & 81.7 \\ \hhline{|=|=|}
		Hovy et al \shortcite{HTH10} -- using WordNet features & 84.8 \\ \hline
		Srikumar and Roth \shortcite{SR13b} -- using WordNet features & 84.78\\ \hline 
		Tratz and Hovy \shortcite{HT09} -- using WordNet features & 76.4 \\ \hline 
		MELB-YB \cite{YB07} -- using WordNet features & 69.3 \\ \hline
		KU \cite{Y07}& 54.7 \\ \hline
		IRST-BP \cite{PTP07} & 49.6 \\ \hline
		Most Frequent Sense  & 39.6 \\ \hline
	\end{tabular}
	}
	\vspace*{-0.2cm}
	\caption{\footnotesize The accuracies on the test set of the SemEval corpus, in comparison to previous systems.} 
	\label{tab:all_semevals}
\end{table}

\subsection{Error Analysis}

\begin{figure*}[ht!]
	\centering
	\subfigure[prepositions]{\label{fig:a}\includegraphics[width=70mm]{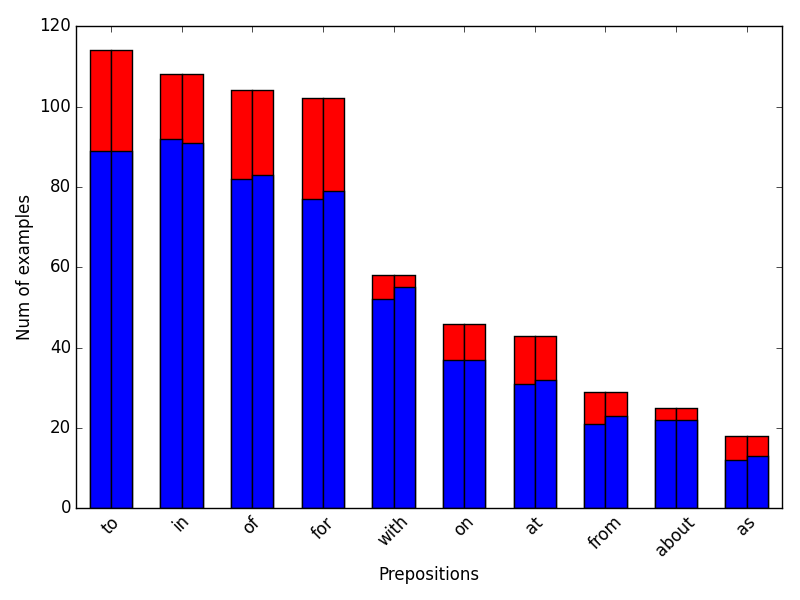}}
	\subfigure[senses]{\label{fig:b}\includegraphics[width=70mm]{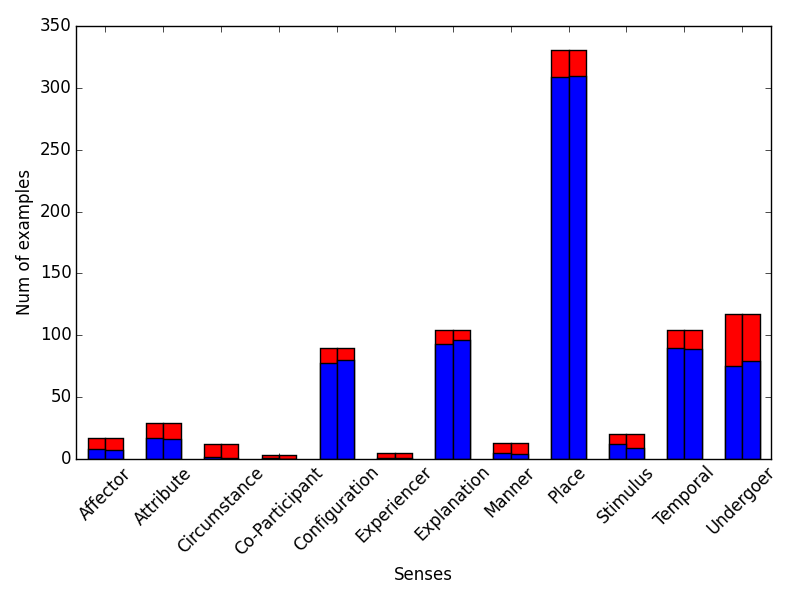}}
	\caption{\footnotesize Assignments on the dev set of the Web-reviews corpus per preposition (a) and per sense (b). Left bars stand for the base model, right bars stand for the multilingual model. In blue are correct assignments, and in red incorrect ones.}
	\label{fig:graph_preps}
\end{figure*}

Figure~\ref{fig:graph_preps} depicts the percentage of correct assignments of the base model, in comparison to the multilingual model, per sense and per preposition (only the 10 most common prepositions are shown). Both models use pre-trained word embeddings and ensembles. Clearly, there is a systematic improvement across most prepositions and senses.

\section{Related work}

\paragraph{Transfer learning and representation learning}

Transfer learning is a methodology that aims to reduce annotation efforts by first learning a model on a different domain or a closely related task, and then transfer the gained knowledge to the main task \cite{PY10}.  
Multi-task learning (MTL) is an approach of transfer learning in which several tasks are trained in parallel while using a shared representation. The different tasks can benefit from each other through this representation \cite{C98}. In this work we use MTL to improve preposition-sense disambiguation, by using an auxiliary multilingual task -- predicting translations of prepositions.

A simple method for sharing information in transfer learning as well as in MTL, is using representations that are shared between related tasks. Representation learning \cite{BCV13} is a closely related field that aims to establish techniques for learning robust and expressive data representations. A well-known effort in this field is that of learning word embeddings for use in a wide range of NLP tasks \cite{MSC13,APS13,LG14,PSM14}. While those representations are highly effective in many cases, other scenarios require representations of a full sentence, or of a context around a target word, rather than representations of single words. Contexts are often represented by some manipulation over the embeddings of their words. Such representations have been successfully used for tasks such as context-sensitive similarity \cite{HSM12}, word sense disambiguation \cite{CLS14} and lexical substitution \cite{MLD15}. An alternative approach for context representation is encoding a context of arbitrary length into a single vector using LSTMs. This approach has been proven to outperform the previous attempts in a variety of tasks such as Semantic Role Labeling \cite{ZX15}, Natural Language Inference \cite{BAP15} and Sentence Completion \cite{MGD16}. We follow the LSTM-based approach for context representation.

\paragraph{Learning from multilingual data}
 
The use of multilingual data for improving monolingual tasks has a long tradition in NLP, and has been used for target word selection \cite{DIS91}; word sense disambiguation \cite{DR02}; and syntactic parsing and named entity recognition \cite{BPB10}, to name a few examples. A dominant approach for exploiting multilingual data is that of cross-lingual projection. This approach assumes a good model exists in one language, and uses annotations in that language in order to constrain possible annotations in another. Projections were successfully used for dependency grammar induction \cite{GGT09}, and for transferring tools such as morphological analyzers and part-of-speech taggers from English to languages with fewer resources \cite {YNW01,YN01}. A different approach is applying multilingual constraints on existing monolingual models, as done for parsing \cite{SS04,BK08} and for morphological segmentation \cite{SB08}.

Of much relevance to this work are also previous attempts to improve monolingual representations using bilingual data \cite{FD14}. Previous works focus on creating sense-specific word embeddings instead of the common word-form specific embeddings \cite{ERC16,STV16}, and also on representing words using their context \cite{KD15,HB13}. While we rely on the assumption most of these works have in common, according to which translations may serve as a strong signal for different senses of words, the novelty of our work is in focusing on prepositions rather than content words, and in jointly representing a context for both a multilingual and a monolingual tasks, which results in an improvement of the monolingual model.

\section{Conclusions and Future Work}

We show that multilingual data can be used to improve the accuracy of preposition-sense disambiguation. The key idea is to train a context-encoder on vast amounts of parallel data, and by that, to obtain a context representation that is predictive of the sense. We show an improvement of the accuracy in all experiments upon using this representation. Our model achieves an accuracy of 80.54 on the Web-reviews corpus, and an accuracy of 81.7 on the SemEval corpus, with significant improvements over models that do not use the multilingual signals. Our result on the SemEval corpus outperforms most previous works, without using any manually curated lexicons.

\section*{Acknowledgements}

The work is supported by The Israeli Science Foundation (grant number 1555/15).

\bibliographystyle{acl}
\bibliography{bib}

\end{document}